\documentclass[conference]{IEEEtran}
\IEEEoverridecommandlockouts
\usepackage{cite}
\usepackage{amsmath,amssymb,amsfonts}
\usepackage{algorithmic}
\usepackage{graphicx}
\usepackage{textcomp}
\usepackage{xcolor}
\usepackage{adjustbox}
\usepackage{booktabs}
\def\BibTeX{{\rm B\kern-.05em{\sc i\kern-.025em b}\kern-.08em
    T\kern-.1667em\lower.7ex\hbox{E}\kern-.125emX}}
\begin{document}

\title{Occlusion-Robust Multi-Object Decoupling for Physics-Based Robotic Interaction
\vspace{-10pt}
\thanks{
% \hrulefill \\
$^{\dagger}$ indicates the corresponding author. 
}
}

\author{Xin Dong$^{1,2}$, Lihan Zhang$^{2}$, Tianru Dai$^{2}$, Wenfeng Deng$^{1,\dagger}$, Yansong Tang$^{2,\dagger}$   \\
$^1$Pengcheng Laboratory $^2$Shenzhen International Graduate School, Tsinghua University \\

}

\maketitle

\begin{abstract}
We propose a mask-free method for lossless multi-object 3D reconstruction from sparse and occluded real-world views, enabling physically plausible robotic interaction via Material Point Method (MPM) simulation. Our key insight is that object coupling stems from occlusion and limited viewpoints, which we address by formulating multi-object decoupling as a sparse-view reconstruction problem. Using 3D Gaussian Splatting as base representation, we first obtain coarse instance partitions with a SAM2–trained segmentation field. Rather than relying on masks, we reconstruct fragmented geometries by leveraging a joint Score Distillation Sampling (SDS) process, which integrates reference-view supervision with novel-view synthesis guided by 2D and 3D diffusion priors to enforce both texture fidelity and 3D consistency. Furthermore, we incorporate geometry-aware priors such as intra-object and inter-object similarity to regularize geometric reasoning. Experimental results demonstrate that our method produces complete, simulation-ready 3D objects without requiring manual masks, enabling realistic dynamic interactions on both synthetic, robotic and real-world datasets.
\end{abstract}


\section{Introduction}
\label{sec:intro}
Interactive 3D reconstruction and physical simulation of realistic scenes play an important role in diverse applications such as industrial digital twin~\cite{robo_twin}, robotic manipulation~\cite{manigaussian,robogs}, virtual reality~\cite{vr-gs,live-gs} and world model~\cite{yu2025wonderworld,physgen,physctrl,animatediff,SAM3D-Phys}. Despite recent efforts~\cite{physgaussian,physdreamer,physflow,omniphysgs,gic,feature_splatting}, performance remains challenging in scenarios involving multi-object coupling. The primary difficulty stems from occlusions and insufficient viewpoint coverage under such coupling conditions, which hinder the non-destructive disentanglement of individual objects during the reconstruction process.

Among existing approaches, PhysGaussian~\cite{physgaussian} treats 3D Gaussians as simulation particles, thereby bridging physics-based simulation tasks with recent advances in 3D Gaussian Splatting~\cite{3dgs}. Methods such as PhysDreamer~\cite{physdreamer}, DreamPhysics~\cite{huang2025dreamphysics}, and PhysFlow~\cite{physflow} leverage knowledge from multimodal large language models and video generation models to infer the material types and physical properties of target objects. Feature Splatting~\cite{feature_splatting} enhances scene understanding by incorporating feature radiance fields, enabling language-guided dynamic interactions. However, all these methods are limited to performing dynamic interactions on a single object within simple scenes. In parallel, several works on amodal completion aim to recover occluded objects. For instance, O2Recon~\cite{O2Recon} employs a 2D image inpainting model to complete missing regions before reconstructing the occluded object in 3D. Amodal3R~\cite{wu2025amodal3r} fine-tunes a 3D generative model to achieve purely 3D amodal reconstruction. DecoupledGaussian~\cite{wang2025decoupledgaussian} attempts to disentangle foreground objects from the background to enable more flexible physics simulation. Nevertheless, none of these approaches can simultaneously address multi-object separation and amodal completion. In contrast, our work tackles the lossless multi-object separation problem specifically tailored for physics simulation. This requires that each separated object be fully recovered in both geometry and texture, ensuring that during simulation, multiple objects can interact accurately and coherently without artifacts such as holes, missing parts, or geometric inconsistencies.

To achieve this,we propose a Mask-Free Multi-Object Decoupling (MF-MOD) method for lossless interactive 3D reconstruction that enables physically robotic interactions through Material Point Method (MPM) simulation. Our key insight is that object coupling primarily stems from occlusion and limited observational viewpoints. Therefore, we reformulate multi-object decoupling as a sparse-view reconstruction problem. Specifically, we build upon 3D Gaussian Splatting~\cite{3dgs} as the base scene representation and obtain instance-level partitions using projected segmentation maps generated by SAM 2. Rather than depending on explicit masks to recover missing geometry, we recover the fragmented parts of objects via a joint Score Distillation Sampling (SDS) process that simultaneously leverages a 2D diffusion prior for novel-view synthesis to ensure high-fidelity textures and a 3D diffusion prior to guide reference-view supervision and enforce global 3D consistency. Furthermore, we introduce geometry-aware priors such as intra-object coherence and inter-object similarity to effectively regularize geometric reasoning. These priors play a crucial role in achieving lossless multi-object decoupling and enabling accurate particle infilling, which are essential for realistic physics-based interaction. 
Our contributions can be summarized as follows:

\begin{figure*}[t]
    \centering
    % \vspace{-15pt}
    \includegraphics[width=1.0\linewidth]{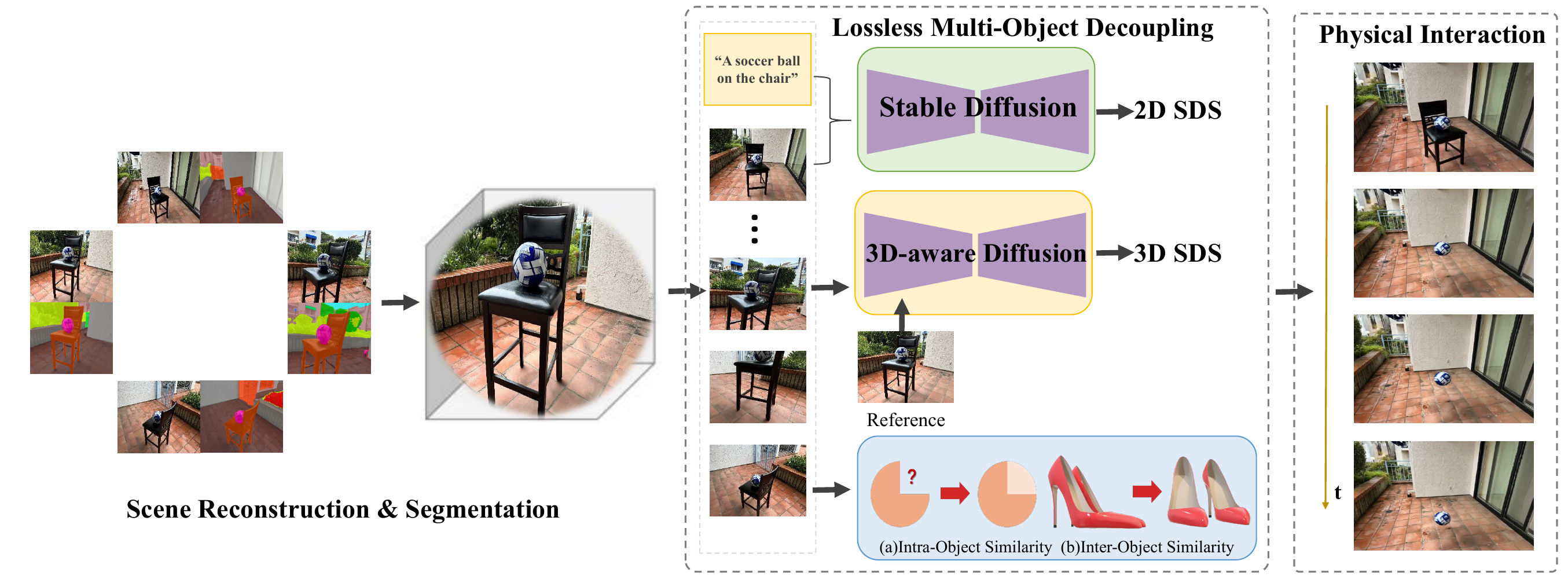}
    \caption{Our framework consists of three main stages: Scene Reconstruction and Segmentation, Lossless Multi-Object Decoupling, and Multi-Object Physical Interaction. Specifically, we utilize joint 2D-3D priors with geometry-aware regularization to achieve geometry recovery and texture inpainting.}
    % \vspace{-5pt}
    \label{fig:framework_1}
\end{figure*}

\begin{itemize}
\item We present a physics-based, interactive 3D reconstruction pipeline for multi-object scenes that supports scene reconstruction, lossless object decoupling, and seamless integration with multi-object robotic interaction.
\item We propose a mask-free multi-object decoupling method that combines joint 2D–3D diffusion priors with geometry-aware regularization to ensure accurate geometry recovery and high-fidelity texture inpainting.
\item Experimental results on both synthetic and real-world datasets demonstrate that our method produces complete, simulation-ready 3D objects without requiring any manual masks, thereby facilitating realistic and dynamic multi-object robotic interactions.
\end{itemize}

\section{Related Work}
\label{sec:rel}

Physical 3D reconstruction and simulation aims to simultaneously reconstruct the visual appearance ($e.g.$, color) of a 3D scene and identify the physical properties of the objects within it, such as material type, elasticity, and hardness. PhysGaussian \cite{physgaussian} leverages 3D Gaussian Splatting as simulation particles for physical dynamics, achieving both high-fidelity rendering and accurate physical simulation. PhysDreamer \cite{physdreamer} exploits dynamic cues from videos to enable the automatic identification of physical attributes. Building upon this, methods like DreamPhysics \cite{huang2025dreamphysics}, Physics3D \cite{physical3d}, and PhysFlow \cite{physflow} further incorporate video generation model priors or optical flow information to extract physical properties more effectively. Additionally, SimAnything \cite{simanything} and Feature Splatting \cite{feature_splatting} utilize multimodal large language models (MLLMs) to achieve zero-shot identification of physical attributes.

However, the aforementioned methods primarily focus on simulating single objects within simplified scenes. They inherently assume that the target object has been completely isolated from the background, an assumption that is rarely valid in real-world scenarios. To address this, DecoupledGaussian \cite{wang2025decoupledgaussian} investigates the extraction of 3D objects from complex real-world scenes to facilitate subsequent physical simulation. Meanwhile, approaches such as IMFine \cite{imfine}, Infusion \cite{infusion}, and Amodal3R \cite{amodal3r} attempt to reconstruct occluded or incomplete objects through 2D inpainting or large-scale 3D dataset training. Nevertheless, none of these methods can achieve lossless separation of entangled multi-object scenes, nor do they address the subsequent reintegration of the extracted objects back into contexts.

\section{Approach}
\label{sec:appr}

\subsection{Overview of Pipeline}

In this section, we present an overview of our pipeline. As illustrated in Fig.~\ref{fig:framework_1}, our framework consists of three main stages: Scene Reconstruction and Segmentation, Lossless Multi-Object Decoupling, and Multi-Object Physical Interaction.
Specifically, in the first stage, given multi-view images and their corresponding segmentation maps extracted using SAM 2, we employ PGSR~\cite{pgsr} to reconstruct the 3D scene. During reconstruction, each Gaussian point is augmented with a segmentation affinity feature that encodes its association with semantic segments. Concurrently, a Multi-Layer Perceptron (MLP) is trained to predict the segmentation label for each Gaussian point based on this feature. This labeling facilitates subsequent object separation and completion. To balance effectiveness and computational efficiency, the dimensionality of the segmentation affinity feature is set to 32.

In the second stage, leveraging the learned segmentation affinity field, we decompose the reconstructed scene into multiple foreground objects and the background, storing them separately. At this point, both objects and background may contain artifacts such as holes or missing regions. To address this, we apply Joint 2D–3D Diffusion Priors together with Reasoning-Based Geometry Regularization to recover the missing geometry of each object. This recovery process ensures geometric completeness and consistent internal infilling, which is important for physically plausible simulation in the next stage.

Finally, the restored objects are reintegrated into the scene, and their dynamic robotic interactions are simulated using a modified Material Point Method (MPM). To accommodate diverse material properties across different objects, we extend the standard MPM formulation by treating each object as a distinct part within a unified simulation entity. This allows simulation particles associated with each part to be assigned unique material parameters, enabling realistic multi-material physical interactions.

\subsection{Joint 2D-3D Priors for Multi-Object Decouling}

Lossless multi-object decoupling involves the geometric and textural completion of objects that exhibit missing regions after separation. We observe that object coupling primarily arises from occlusion and limited viewing angles. Consequently, this challenge can be effectively addressed by modeling multi-object decoupling as a sparse-view reconstruction problem. Specifically, inspired by recent advances in single-image-to-3D generation, we move beyond traditional mask-based approaches and instead recover the missing portions of fragmented geometries through a unified Score Distillation Sampling (SDS) framework. This framework integrates two complementary priors: reference-view supervision guided by a 2D diffusion prior to ensure high-fidelity texture reconstruction, and novel-view synthesis guided by a 3D diffusion prior to enforce global 3D consistency.

\begin{figure*}[t]
    \centering
    % \vspace{-15pt}
    \includegraphics[width=1.0\linewidth]{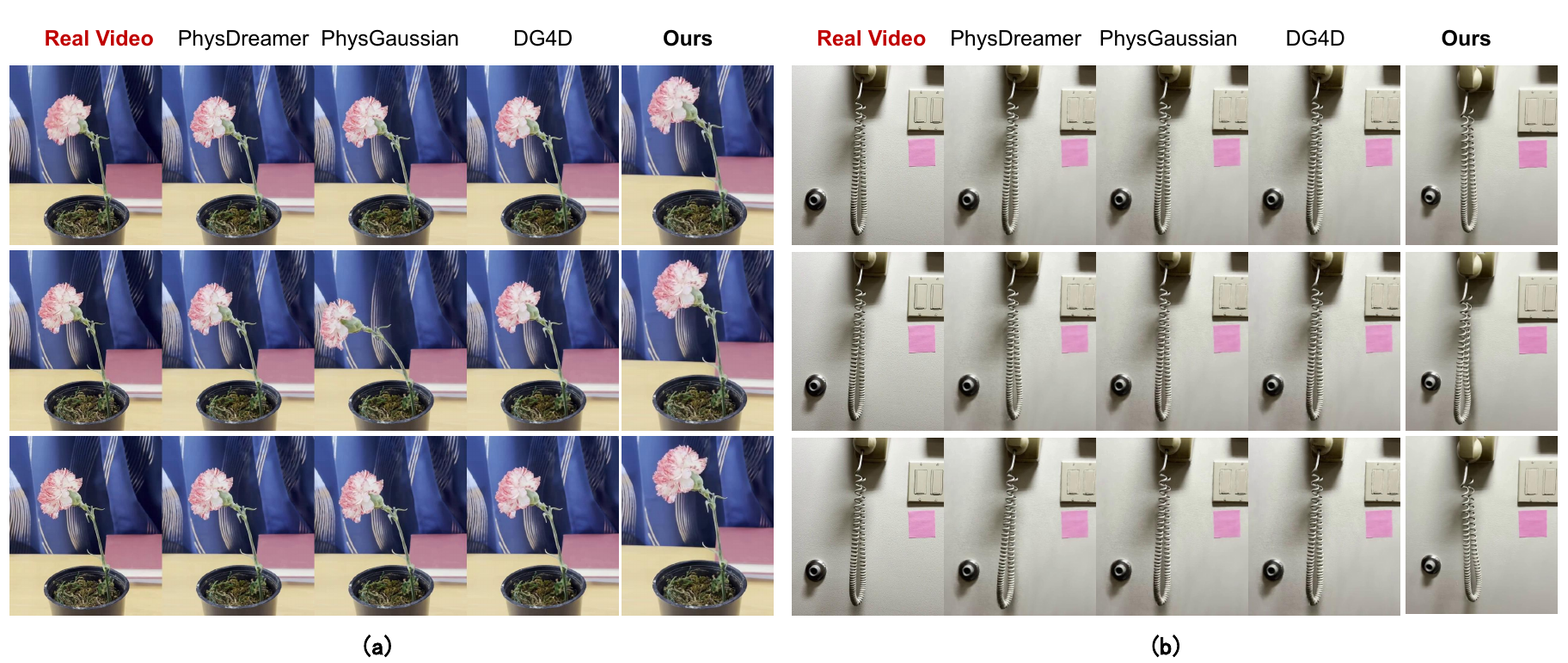}
    \caption{We compare motion effects after robotic interaction across different objects. The dynamic scenes are rendered as videos for visualization, with three uniformly sampled frames shown for each.}
    % \vspace{-5pt}
    \label{fig:exp1}
\end{figure*}

\textbf{2D Diffusion Prior. }
We employ Score Distillation Sampling (SDS)~\cite{SDS} as the 2D diffusion prior for object recovery. SDS serves as a pivotal mechanism for distilling knowledge from pre-trained 2D diffusion models into 3D scene generation, facilitating sparse-view 3D synthesis without explicit 3D supervision. By exploiting the score ($i.e.$, the gradient of the data log-likelihood) predicted by the 2D diffusion model on rendered images, SDS directs the optimization of 3D Gaussian splatting parameters, with particular emphasis on occluded or unobserved regions. Specifically, given a differentiable renderer that maps 3D scene parameters $\boldsymbol{\theta}$ to an image $\mathbf{I} = \mathcal{M}(\boldsymbol{\theta})$, SDS computes the gradient of a pseudo-loss in the 3D space via the chain rule. The loss function is defined as:
$$
\mathcal{L}_{2D-SDS} = \mathbb{E}_{t,\epsilon} \left[ w(t) \left( \epsilon_\phi(\mathbf{z}_t; \mathbf{l}, t) - \epsilon \right) \frac{\partial \mathbf{z}}{\partial \mathbf{I}} \frac{\partial \mathbf{I}}{\partial \boldsymbol{\theta}} \right],
$$
where $\mathbf{z}_t$ is the noisy version of the rendered image at timestep $t$, $\epsilon_\phi$ is the noise predictor of the 2D diffusion model, $\epsilon \sim \mathcal{N}(0, I)$ is the true noise, $\mathbf{l}$ denotes the conditioning input prompt, and $w(t)$ is a time-dependent weighting function. The term $\frac{\partial \mathbf{z}}{\partial \mathbf{I}}$ represents the gradient of the loss with respect to the image, and $\frac{\partial \mathbf{I}}{\partial \boldsymbol{\theta}}$ is the Jacobian of the rendering process, which backpropagates the 2D signal into the 3D parameter space. This formulation allows SDS to optimize 3D representations so that their rendered views are consistent with the semantic and visual priors encoded in the 2D diffusion model.

\textbf{3D-aware Diffusion Prior. }
$L_{2D-SDS}$ enables the generation of textured 3D content without requiring explicit 3D supervision. However, it does not guarantee 3D consistency, $i.e.$ the appearance of the same 3D object may vary across different viewpoints. This limitation arises because the 2D SDS loss enforces gradients of rendered images from arbitrary views to align with textual guidance, without explicitly modeling inter-view relationships. 

To address this issue, we draw inspiration from~\cite{Magic123} and modify the SDS loss. Instead of directly pulling the gradient of each rendered view toward the text-conditioned target, we align it with the rendering from a designated reference view. This modification offers two key advantages. First, it endows the model with an implicit understanding of 3D structure, thereby ensuring visual consistency across multiple viewpoints. Second, by leveraging a complete reference view as a geometric and textural prior, it effectively recovers missing regions in other views, leading to more coherent and complete 3D geometry and texture. The modified loss function can be formulated as follows:
\begin{equation}
    \mathcal{L}_{3D-SDS} = \mathbb{E}_{t,\epsilon} \left[ w(t) \big( \epsilon_\phi(\mathbf{z}_t; \mathbf{I}^r, t, R, T) - \epsilon \big) \frac{\partial \mathbf{I}}{\partial \theta} \right],
    \label{eq:3d_loss}
\end{equation}
here, $\mathbf{I}^r$ is the reference view, and $R, T$ are the camera poses of it. In our implementation, we adopt the pre-trained model weights from~\cite{Zero123}, which were trained on a dataset comprising over 800,000 pairs.  

Overall, by combining 2D Score Distillation Sampling (SDS) with our 3D-aware SDS formulation, our approach ensures high-fidelity texture reconstruction while enforcing global 3D consistency across multiple viewpoints. Thus, the total reconstruction loss is:
\begin{equation}
    L_{all} = L_{color} + \lambda_1 L_{2D-SDS} + \lambda_2 L_{3D-SDS},
    \label{eq:loss_total}
\end{equation} 
where $L_{color}$ is the rgb loss from original 3D gaussian splatting training process. $\lambda_1$ and $\lambda_1$ are set to 1e-4.

\begin{figure*}[t]
    \centering
    % \vspace{-15pt}
    \includegraphics[width=0.85\linewidth]{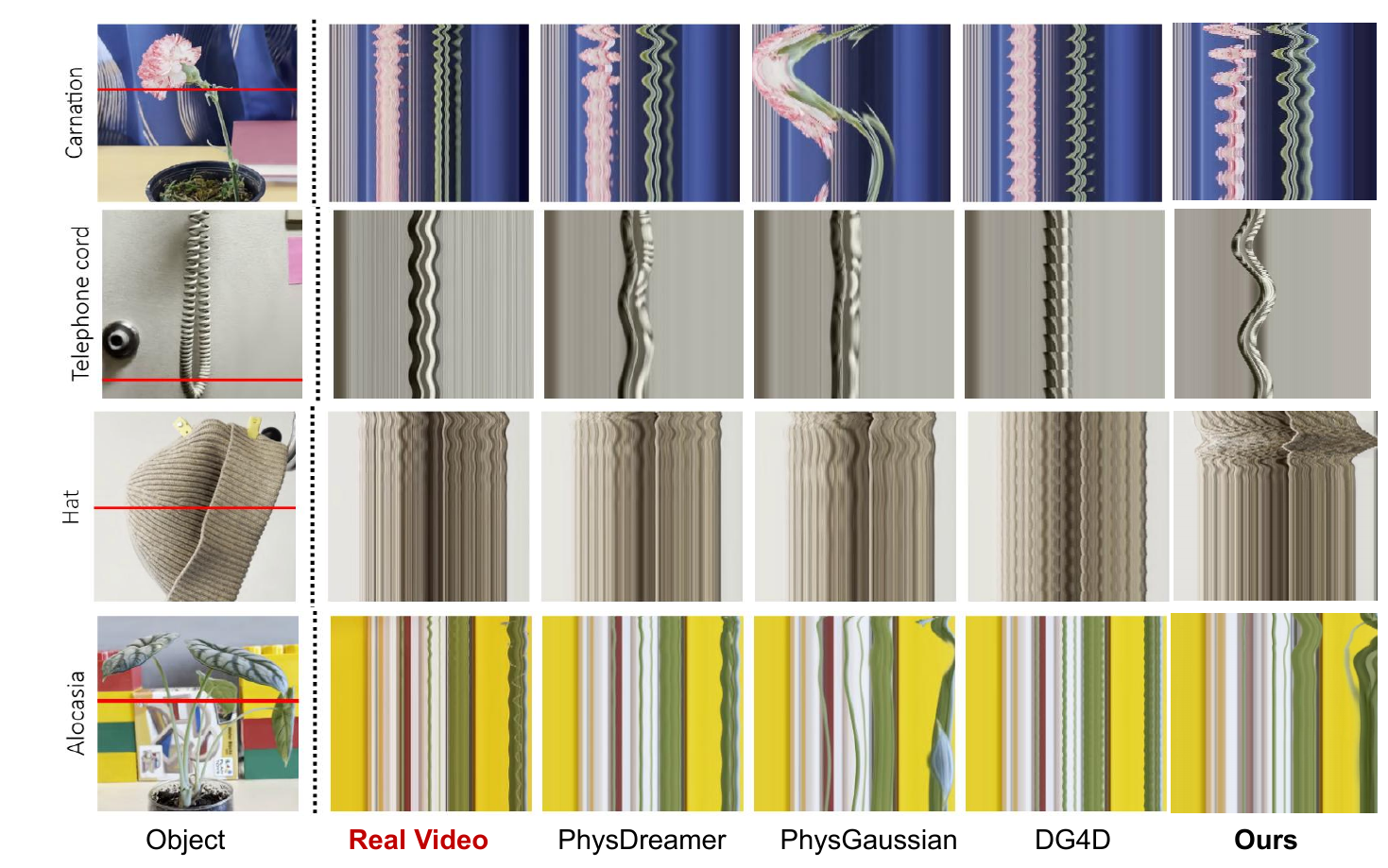}
    \caption{To better highlight temporal consistency and physical plausibility after robotic interaction, we present time-slice visualizations that combine outputs generated at different timesteps from the same spatial viewpoint.}
    % \vspace{-5pt}
    \label{fig:exp2}
\end{figure*}

\subsection{Similarity-Based Geometry Reasoning}

To mitigate the geometric and appearance coupling induced by occlusion and insufficient observations among neighboring or contacting objects, we further introduce a similarity-based geometry reasoning strategy for decoupling regularization. This strategy exploits two complementary structural priors: (1) intra-object self-similarity, which reflects the repeatability and consistency of local geometric structures within the same object across varying viewpoints; and (2) inter-object mutual similarity, which captures the commonalities in shape, material among multiple instances of the similar object category. The schematic diagram is shown in Fig.~\ref{fig:framework_1}.

Consider a scene comprising \( N \) objects that exhibit regular geometric structures and belong to similar semantic categories. Partial observations of these objects are provided in the form of color and segmentation maps, denoted as \( \{ \mathbf{X}_i^{\text{obs}} \}_{i=1}^N \), where each \( \mathbf{X}_i^{\text{obs}} \subset \mathbb{R}^{3 \times M_i} \) represents the incomplete observation of the \( i \)-th object, with \( M_i \) being the number of observed points for that object. We define the inter-object mutual similarity guided completion function $ \mathcal{F}_{\text{recover}} $ that aims to recover the complete geometry and texture $\mathbf{X}_i^{\text{full}}$:
\begin{equation}
\label{total}
    \mathbf{X}_i^{\text{full}} = \mathcal{F}_{\text{recover}} \left( \mathbf{X}_i^{\text{obs}}, \, \{ \mathbf{X}_j^{\text{obs}} \}_{j \neq i}, \, \Phi_{\text{sim}}, \mathbf{L} \right),
\end{equation}
where $ \Phi_{\text{sim}} $ represents semantic and structural priors provided by a multimodal foundation model to guide similarity matching. $L$ is the textual description of objects, which can use the reasoning ability of multi-modal foundation model. Specifically, inter-object mutual similarity is realized through cross-object feature alignment:
\begin{equation}
\label{simi_compute}
    \mathbf{S}_{ij} = \mathrm{Sim}\left( \mathcal{E}(\mathbf{X}_i^{\text{obs}}), \, \mathcal{E}(\mathbf{X}_j^{\text{obs}}) \right),
\end{equation}
where $ \mathcal{E}(\cdot) $ is a learnable geometric encoder, and $ \mathrm{Sim}(\cdot,\cdot) $ denotes a similarity metric, $i.e.$ cosine similarity. The missing regions are then reconstructed via weighted Gaussian particle aggregation from reference regions:
\begin{equation}
\label{missing_equ}
    \mathbf{X}_i^{\text{missing}} = \sum_{j \neq i} w_{ij} \cdot \mathcal{T}_{j \rightarrow i} \left( \mathbf{X}_j^{\text{obs}} \right),
\end{equation}
here $w_{ij} = \frac{\exp(\mathbf{S}_{ij})}{\sum_{k \neq i} \exp(\mathbf{S}_{ik})}$,and $ \mathcal{T}_{j \rightarrow i} $ denotes a geometric alignment transformation ($i.e.$ aggregation of weighted Gaussian particles from reference regions) from object $ j $ to object \( i \), determined jointly by pose, scale, and semantic context. Note that when we generalize the notion of an object to encompass arbitrary regions and allow indices i and j to represent any such regions, Equ.~\ref{total}, Equ.~\ref{simi_compute} and Equ.~\ref{missing_equ} can be expressed as an intra-object self-similarity-guided completion function that captures the relationships between distinct regions within the same object.

Through this strategy, our model can infer missing geometry in a physically plausible manner without relying on explicit annotations and masks. This effectively disentangles the appearance–geometry coupling inherent in scenes with multiple interacting objects while preserving the geometric and semantic integrity of each individual instance. Consequently, it yields a lossless and physically consistent initial state that is well suited for subsequent dynamic interaction simulation. The approach is especially effective for object categories characterized by strong structural regularities, such as furniture and industrial components.

\section{Experiments}
\label{sec:appr}

In this section, we first present the experimental settings, including the datasets used, implementation details, and evaluation metrics. We then compare our method against state-of-the-art approaches to highlight its advantages. Finally, we provide ablation studies to demonstrate the effectiveness of each individual module.

\subsection{Experimental Settings}

\textbf{Datasets. } For a fair and consistent comparison, we select four real-world static scenes from PhysDreamer~\cite{physdreamer}. The selected objects include a carnation, an alocasia plant, a coiled telephone cord, and a beanie hat. In addition, we incorporate the soccer ball asset provided in Feature Splatting~\cite{feature_splatting} and supplement it with a tissue box scene that we captured ourselves. These real-world assets validate the performance of our method in practical scenarios.

\textbf{Metrics. } For real-world scenes lacking ground-truth material properties, we evaluate our method using visual quality metrics of rendered videos as proxies. We adopt three standard metrics: PSNR, which measures pixel-wise fidelity; SSIM, which assesses similarity in luminance, contrast, and structural information; and MS-SSIM, an extension that evaluates structural consistency across multiple scales for a more holistic assessment of perceptual quality.

\subsection{Comparison with the State of the Art}

In this subsection, we conduct a comparative evaluation against state-of-the-art methods, focusing first on the visual quality and dynamic patterns of generated scenes, and then on our method’s ability to recover both individual objects and full scenes under multi-object decoupling.

As illustrated in Fig.~\ref{fig:exp1}, we compare motion effects after robotic interaction across different objects. The synthesized dynamic scenes are rendered as videos, with three uniformly sampled frames shown for each. To better highlight temporal consistency and physical plausibility, Fig.~\ref{fig:exp2} presents time-slice visualizations that combine outputs generated at different timesteps from the same spatial viewpoint. These results demonstrate that our method produces dynamics consistent with real-world physics, whereas competing approaches exhibit limitations. For example, DG4D is restricted to periodic motions, and PhysGaussian generates unnatural or physically implausible deformations.

Additionally, Table~\ref{tab:comp_tab} reports quantitative comparisons of visual quality in dynamic scene synthesis. We compare our method with PhysDreamer~\cite{physdreamer}, PhysGuassian~\cite{physgaussian} and Physics3D~\cite{Liu2024Physics3DLP} Our method achieves visual fidelity comparable to the best-performing approaches while simultaneously preserving physically accurate dynamics.

\begin{table}[t]
    \small
    \centering
    \caption{Quantitative comparison of our method against recent approaches on visual metrics.}
    \begin{adjustbox}{max width=\textwidth}
    \begin{tabular}{lccc}
        \toprule
         & PSNR$\uparrow$ & SSIM$\uparrow$ & MS-SSIM$\uparrow$ \\
        \midrule
        PhysDreamer & 13.89 & 0.55 & 0.37 \\
        PhysGaussian & 13.86 & 0.57 & 0.39 \\
        Physics3D & 14.72 & 0.59 & \textbf{0.49} \\
        Ours & \textbf{17.83} & \textbf{0.59} & 0.45 \\
        \bottomrule
    \end{tabular}
    \end{adjustbox}
    \vspace{-10pt}
    \label{tab:comp_tab}
\end{table}

\begin{table}[t]
    \vspace{-10pt}
    \small
    \centering
    \caption{Ablation studies isolating the contribution of each module and ablating the choice of segmentation model (SAM vs. SAM2). SBGR denotes similarity-based geometry reasoning.}
    \begin{adjustbox}{max width=\textwidth}
    \begin{tabular}{lccc}
        \toprule
         & PSNR$\uparrow$ & SSIM$\uparrow$ & MS-SSIM$\uparrow$ \\
        \midrule
        w/o SBGR & 10.21 & 0.32 & 0.30 \\
        w/o 3D-aware Loss & 16.57 & 0.47 & 0.41 \\
        w/o 2D-SDS & 14.28 & 0.39 & 0.31 \\
        MF-MOD & \textbf{17.83} & \textbf{0.59} & \textbf{0.45} \\
        \midrule
        MF-MOD (SAM) & 17.63 & 0.52 & 0.43 \\
        MF-MOD (SAM2) & \textbf{17.83} & \textbf{0.59} & \textbf{0.45} \\
        \bottomrule
    \end{tabular}
    \end{adjustbox}
    \vspace{-5pt}
    \label{tab:abl_tab}
\end{table}

\begin{figure}[t]
    \centering
    % \vspace{-15pt}
    \includegraphics[width=\linewidth]{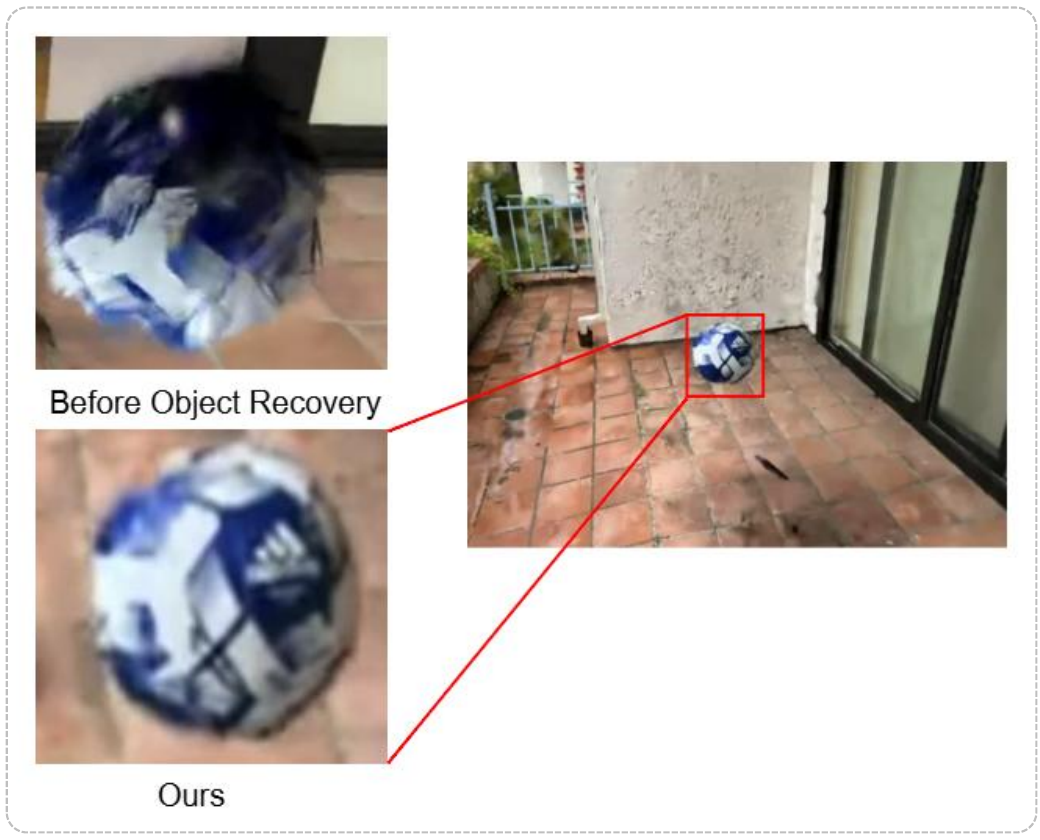}
    \caption{Prior to applying our method, the back side of the soccer ball shows clear holes and jagged artifacts. Our approach, which combines joint 2D–3D diffusion with a geometric similarity prior, effectively infers the geometry and texture of the occluded regions.}
    \vspace{-2pt}
    \label{fig:exp3}
\end{figure}

\subsection{Ablation Studies}
We conducted ablation studies to assess each module’s contribution. As shown in the upper part of Table~\ref{tab:abl_tab}, removing any module degrades performance. Notably, omitting similarity-based geometry reasoning  distorts geometry, drastically worsening all metrics. We also compared segmentation models (SAM vs. SAM2) in the lower part of Table~\ref{tab:abl_tab}, SAM2 offers slight gains, but the difference is minor, confirming our method’s effectiveness and insensitivity to the segmentation model version.

\subsection{Evaluation of Decoupling and Robotic Interaction}

To validate the effectiveness of our method in multi-object decoupling, particularly its ability to recover both individual objects and full scenes, we present results on the soccer ball example in Fig.~\ref{fig:exp3}. Prior to applying our method, the back side of the soccer ball shows clear holes and jagged artifacts. Our approach, which combines joint 2D–3D diffusion with a geometric similarity prior, effectively infers the geometry and texture of the occluded regions.

\begin{figure}[t]
    \centering
    % \vspace{-15pt}
    \includegraphics[width=1.0\linewidth]{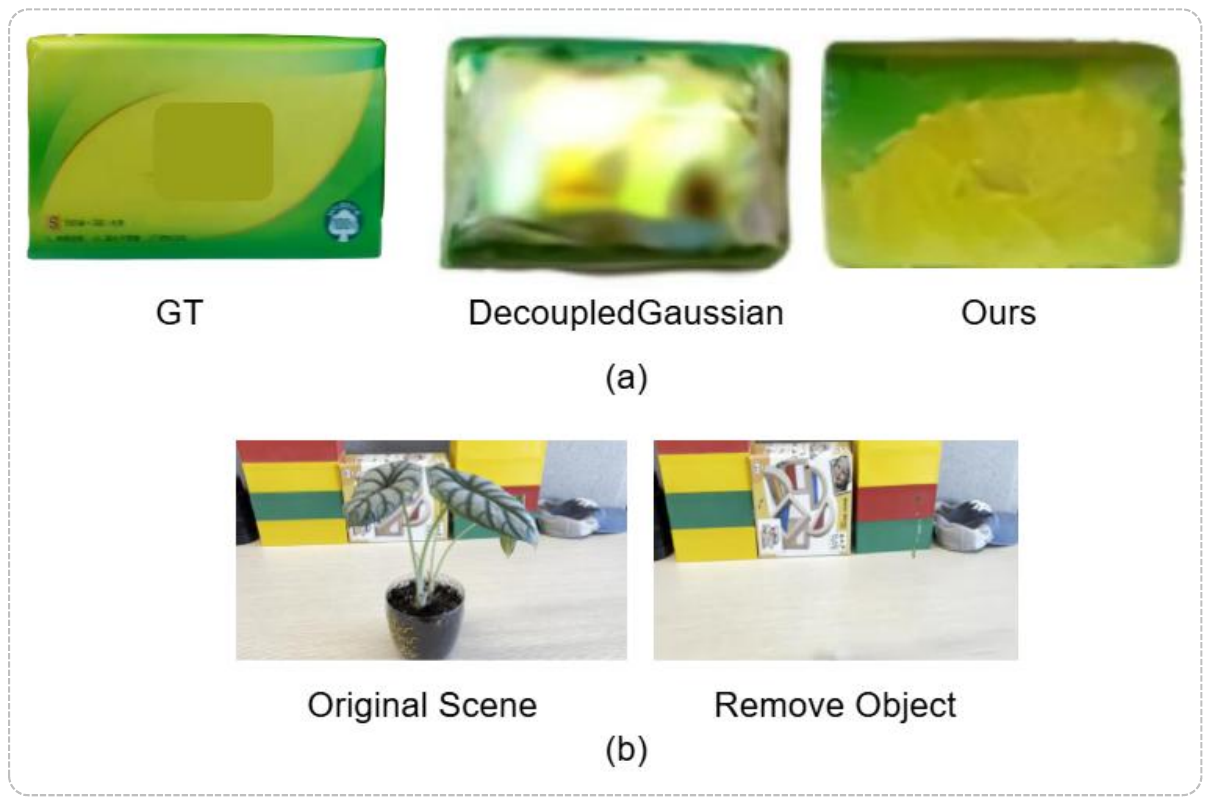}
    \caption{To validate the effectiveness of our method in multi-object decoupling, particularly its ability to recover both individual objects and full scenes, we present results on the soccer ball and tissue box examples.}
    % \vspace{-5pt}
    \label{fig:exp4}
\end{figure}

As shown in Fig.~\ref{fig:exp4}(a), we evaluate completion performance on a more complex everyday object: a tissue box. Compared to DecoupledGaussian, our method achieves higher geometric accuracy and substantially improved texture quality. In addition, Fig.~\ref{fig:exp4}(b) demonstrates our scene recovery capability. When the foreground object (an alocasia plant) is removed, our method successfully reconstructs the background, indicating its ability to maintain scene consistency even in the absence of foreground elements. These results collectively confirm the effectiveness of our approach in multi-object decoupling and holistic scene restoration.

\begin{figure}[t]
    \centering
    % \vspace{-15pt}
    \includegraphics[width=1.0\linewidth]{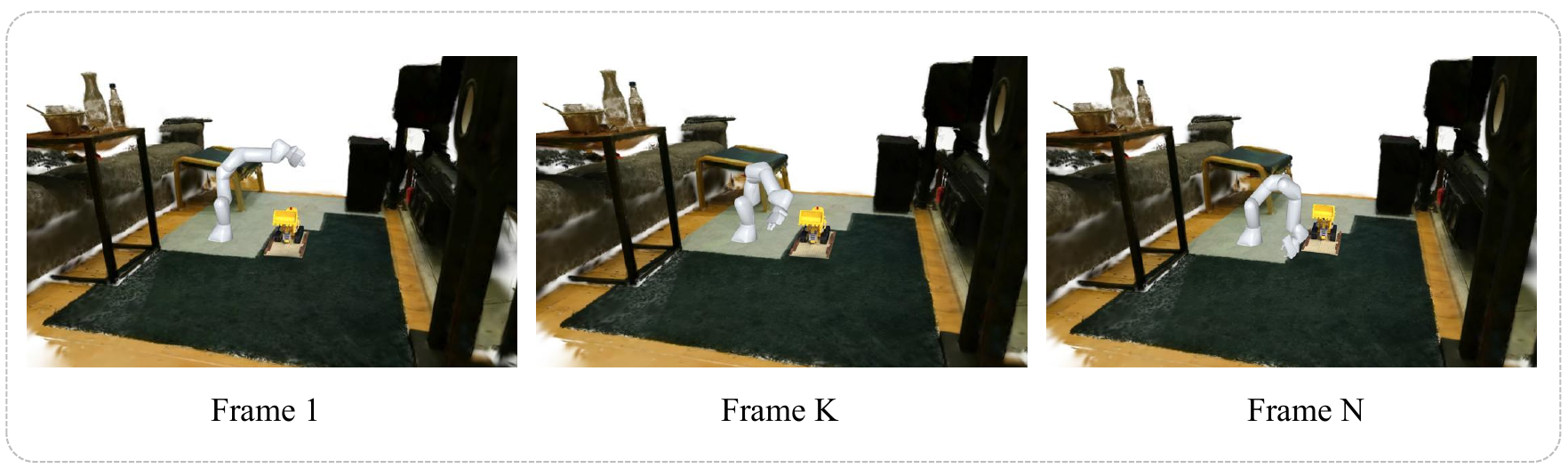}
    \caption{To validate our method's efficacy in industrial robotic interaction, we show the precise collision interactions between robotic arm and extracted 3D object. The first, second, and third images present the initial frame, the $K$-th frame, and the final frame, respectively, all rendered from the dynamic 3D robotic interaction.}
    % \vspace{-5pt}
    \label{fig:robotic_interaction}
\end{figure}

Moreover, to demonstrate our method's efficacy in industrial robotic interaction, we integrated a Franka robotic arm into a room scene containing objects extracted and completed from a toy dozer scene. As shown in Fig.~\ref{fig:robotic_interaction}, the resulting precise collision interactions confirm that our reconstructed 3D objects possess substantial practical value for industrial robotic manipulation tasks.

\section{Conclusion}

We propose Mask-Free Multi-Object Decoupling (MF-MOD), a mask-free method for lossless interactive 3D reconstruction that enables physically plausible robotic interactions via Material Point Method (MPM) simulation. By formulating multi-object decoupling as a sparse-view reconstruction problem, MF-MOD recovers fragmented geometries through a unified Score Distillation Sampling (SDS) that combines 2D and 3D diffusion priors. Moreover, similarity-based geometry reasoning capturing intra-object and inter-object similarity further regularizes the recovery process. Experiments demonstrate that MF-MOD produces complete, simulation-ready 3D objects without manual masks, enabling realistic dynamic interactions on both synthetic, robotic and real-world datasets.

\textbf{Limitations. } Our approach processes multi-object decoupling in a sequential manner, which incurs substantial computational and temporal overhead when dealing with a large number of objects. Furthermore, while our method infers missing regions based on visible areas, it does not account for color variations caused by lighting differences when objects are separated. We will address these two limitations in our future work.

\bibliographystyle{IEEEbib}
\bibliography{reference}

\end{document}